\documentclass{article}
\usepackage{stywhispers,amsmath,epsfig}
\usepackage{graphicx}
\usepackage{color,soul}
\usepackage{hyperref}
\usepackage{mwe}
\usepackage{subcaption}
\usepackage{amsmath}
\usepackage{tabularx} 
\usepackage{multirow}
\usepackage{booktabs}

\newcolumntype{L}{>{\centering\arraybackslash}m{3cm}}
\usepackage{float}
\usepackage[thinc]{esdiff}
\usepackage{graphicx} 
\usepackage[table]{xcolor}
\usepackage[T1]{fontenc}
\usepackage{caption}
\usepackage{enumitem}
\setlist[itemize]{noitemsep, topsep=0pt}
\newlength{\bibitemsep}
\newlength{\bibparskip}
\let\oldthebibliography\thebibliography
\renewcommand\thebibliography[1]{%
  \oldthebibliography{#1}%
  \setlength{\parskip}{\bibitemsep}%
  \setlength{\itemsep}{\bibparskip}%
}

\title{Hyper-Drive: Visible-Short Wave Infrared Hyperspectral Imaging Datasets for Robots in Unstructured Environments}

\name{Nathaniel Hanson$^{1*}$, Benjamin Pyatski$^{1}$, Samuel Hibbard$^{1}$, Charles DiMarzio$^{2}$, Taşkın Padır$^{1}$
\thanks{Research was sponsored by the United States Army Core of Engineers (USACE) Engineer Research and Development Center (ERDC) Geospatial Research Laboratory (GRL) and was accomplished under Cooperative Agreement Federal Award Identification Number (FAIN) W9132V-22-2-0001. The views and conclusions contained in this document are those of the authors and should not be interpreted as representing the official policies, either expressed or implied, of USACE EDRC GRL or the U.S. Government. The U.S. Government is authorized to reproduce and distribute reprints for Government purposes notwithstanding any copyright notation herein.}
\thanks{*Corresponding author {\tt\small hanson.n@northeastern.edu}}}
\address{$^{1}$Institute for Experiential Robotics; $^{2}$Electrical and Computer Engineering Department\\Northeastern University, Boston, Massachusetts, USA}
\begin{document}
%
\maketitle
\begin{abstract}

Hyperspectral sensors have enjoyed widespread use in the realm of remote sensing; however, they must be adapted to a format in which they can be operated onboard mobile robots. In this work, we introduce a first-of-its-kind system architecture with snapshot hyperspectral cameras and point spectrometers to efficiently generate composite datacubes from a robotic base. Our system collects and registers datacubes spanning the visible to shortwave infrared (660-1700 nm) spectrum while simultaneously capturing the ambient solar spectrum reflected off a white reference tile. We collect and disseminate a large dataset of more than 500 labeled datacubes from on-road and off-road terrain compliant with the ATLAS ontology to further the integration and demonstration of hyperspectral imaging (HSI) as beneficial in terrain class separability. Our analysis of this data demonstrates that HSI is a significant opportunity to increase understanding of scene composition from a robot-centric context. All code and data are open source online: \url{https://river-lab.github.io/hyper_drive_data}
\end{abstract}
\section{Introduction}
\label{sec:intro}
Mirroring human-level terrain perception in robots is area of active research, given its criticality in enabling actionable intelligence prior to traversing the surface. Terrain can be best understood as a type of abstract material of unspecified extent \cite{adelson2001seeing}.  Unlike objects defined by regular geometric properties, terrain challenges traditional perception systems because surface materials vary widely in their size and shape. For instance, a farm field and patch of soil have the same texture and micro appearance, but cover drastically different geographic areas. Coarse labels like grass, soil, and sand are useful in semantic segmentation, but intra-class differences affect traversability. On road surfaces, these features might manifest as oil slicks, standing water, or black ice. 

\begin{figure}[t]
    \includegraphics[width=\linewidth]{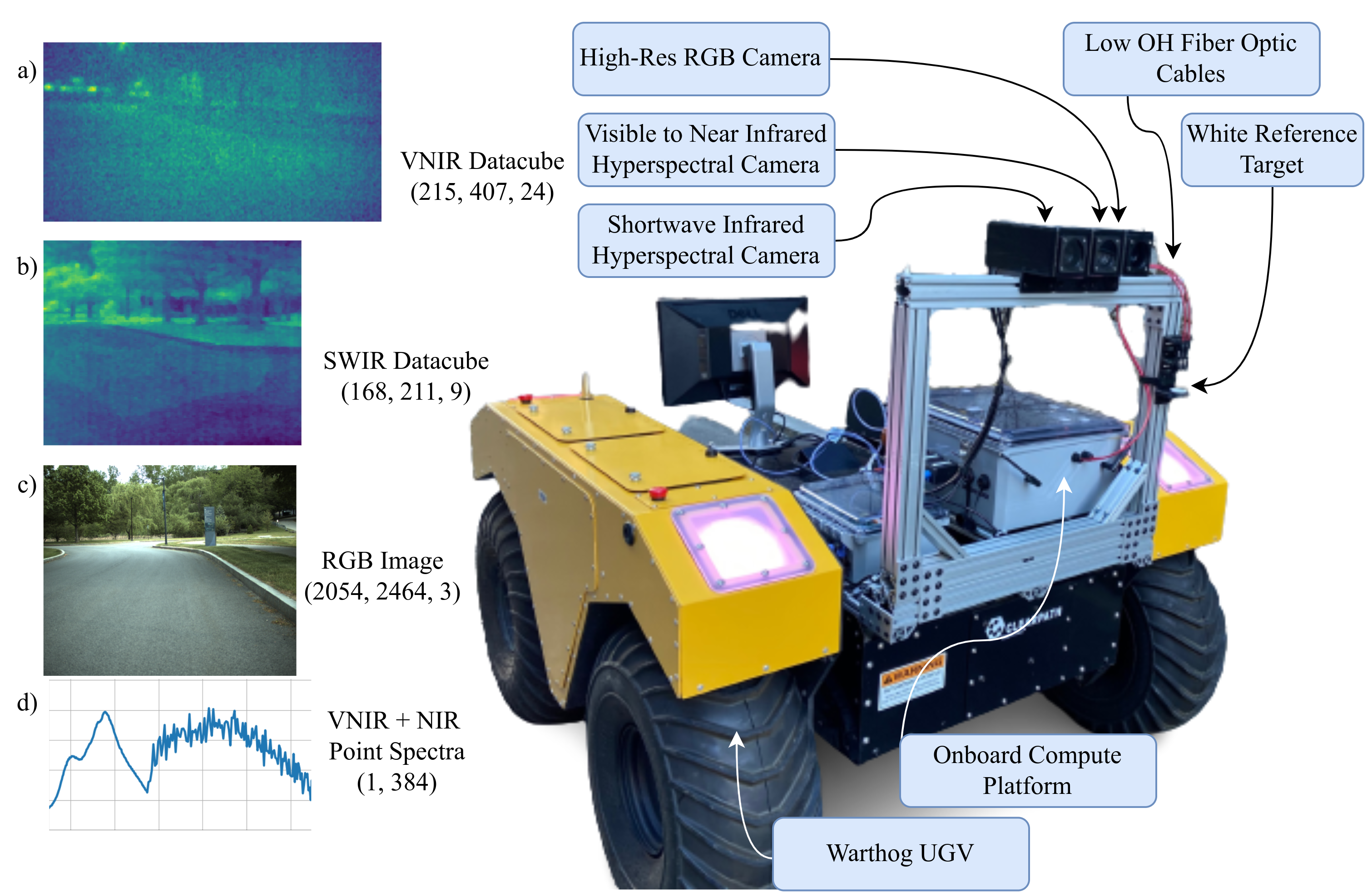}
    \caption{Hyper-Drive system mounted to off-road mobile robot, with sample data representations of white reference target from a) the Visible to Near Infrared (VNIR) hyperspectral camera b) Shortwave Infrared hyperspectral camera c) High resolution RGB camera d) Combined point spectrometers.}
    \label{fig:hyper_drive_teaser}
    \vspace{-1.5em}
\end{figure}

Our approach to this problem builds on our previous work in terrain classification with point-based spectroscopy and multi-modal sensing \cite{hanson2022vast}. This present work introduces hyperspectral imaging (HSI) to mobile robot terrain understanding. In this research, we develop a system, HYPER DRIVE, shown in Fig.~\ref{fig:hyper_drive_teaser}, to capture datacubes from a moving platform with variable illumination. Our work shows forward-facing HSI is a powerful tool for robotics and is useful in a variety of terrain conditions. The contributions of this paper are as follows.
\begin{itemize}
    \item Development of a sensing system to incorporate Visible to Short Wave Infrared hyperspectral cameras onto a mobile robot with solar reference spectrum.
    \item Open-source software framework and message types implemented for the Robot Operating System \cite{quigley2009ros}
    \item Multi-modal dataset containing registered imaging products and ambient solar spectra data for system.
\end{itemize}
\section{RELATED WORK}
\label{sec:prior}
Hyperspectral data acquired from moving vehicles is still in its infancy. The recent innovation of snapshot hyperspectral cameras with fast integration times has made it possible to capture images of a nonplanar scene, while either the camera or objects are moving relative to each other. The absence of artifacts like motion blurring is critical to obtain representative spectra and distinct spatial characteristics. 

\subsection{Hyperspectral Terrain Datasets}
There have been multiple efforts to develop vehicle-mounted hyperspectral cameras to collect datacubes from off-road \cite{jakubczyk2022hyperspectral, winkens2017hyko}, on-road \cite{basterretxea2021hsi, you2019hyperspectral, huang2021weakly}. Notably, all the aforementioned examples make use of Visible-Near Infrared (VNIR) cameras, which are well-suited to detect vegetative properties, but do not have the same insight into the numerous absorption bands in the shortwave infrared spectrum \cite{gupta2003comparative}. Unlike RGB image terrain segmentations which have adopted standard label sets such as the widely-used KITTI classes \cite{geiger2013vision}, HSI datasets are contextually driven with labels largely chosen by the operating range of the camera. Examples of semantic label sets from the literature include (dataset name bolded):
\begin{itemize}
    \item \textbf{HSI-Drive}: Road, road marks, vegetation, painted metal, sky, concrete/stone/brick, pedestrian/cyclist, water, unpainted metal, glass/transparent plastic \cite{basterretxea2021hsi}.
    \item \textbf{Winkens et al.} Drivable, rough, obstacle, sky\cite{winkens2017hyperspectral}
    \item \textbf{HyKo} Plastic, soil, paper, street, wood, chlorophyll, metal, sky \cite{winkens2017hyko}.
    \item \textbf{Jakubczyk et al.} Ground road, forest road, asphalt road, grass, forest \cite{jakubczyk2022hyperspectral}.
    \item \textbf{Hyperspectral City V1.0}: Car, human, road, traffic light, traffic sign, tree, building, sky, object \cite{you2019hyperspectral}.
    \item \textbf{HSI Road}: Road, background \cite{lu2020hsi}.
\end{itemize}

\cite{basterretxea2021hsi} contains annotations for the time of day and the time of year and aggregations of classes into categories including drivable / non-drivable, drivable / road markers / vehicles and drivable / road markers / vehicles / pedestrians \cite{basterretxea2021hsi}. Similarly \cite{winkens2017hyko} included multi-modal sensors including a VNIR spectrometer and Light Detection and Ranging (LIDAR) sensors in their HyKo dataset, but did not elaborate on how the spectrometer could benefit the system calibration \cite{winkens2017hyko}. HyKo also contains condensed annotations on drivability classes (rough, sky, obstacle, drivable).

\subsection{Spectral Informed Terrain Understanding}
\cite{winkens2017hyperspectral} shows the potential of HSI in terrain classification, even with simple machine learning methods such as random forest, and later through manual feature extraction \cite{winkens2017robust}.

\cite{huang2021weakly} claims HSI overcomes challenges induced by object metamerism. RGB images are empirically shown to have a lower degree of separability than HSI datacubes. The authors exploit this information in a semantic segmentation network, with a finetuning module leveraging 10 classes that combine semantic purpose and material information. 

\cite{cai2022rgb} explores the use of RGB imagery alone to perform material segmentation with transform-based neural networks. They also propose a dataset called KITTI-Materials, containing 1000 frames with 20 different material categories \cite{cai2022rgb}. The same group also augmented RGB imagery with NIR and polarization images to improve classification accuracy on materials such as metal and water \cite{liang2022multimodal}.

In our previous work, we demonstrated high accuracy in terrain classification by measuring the spectral signatures of terrain near the contact point on wheeled vehicles \cite{hanson2022vast}. Combining feature-specific neural networks for IMU and RGB image classification through a fusion network provided increased performance. The results of \cite{ilehag2020urban} show that further fusing visual imagery and NIR spectral signatures aids in the classification of urban road surface types.

\begin{figure*}[t]
    \vspace{0.5em}
    \includegraphics[width=\linewidth]{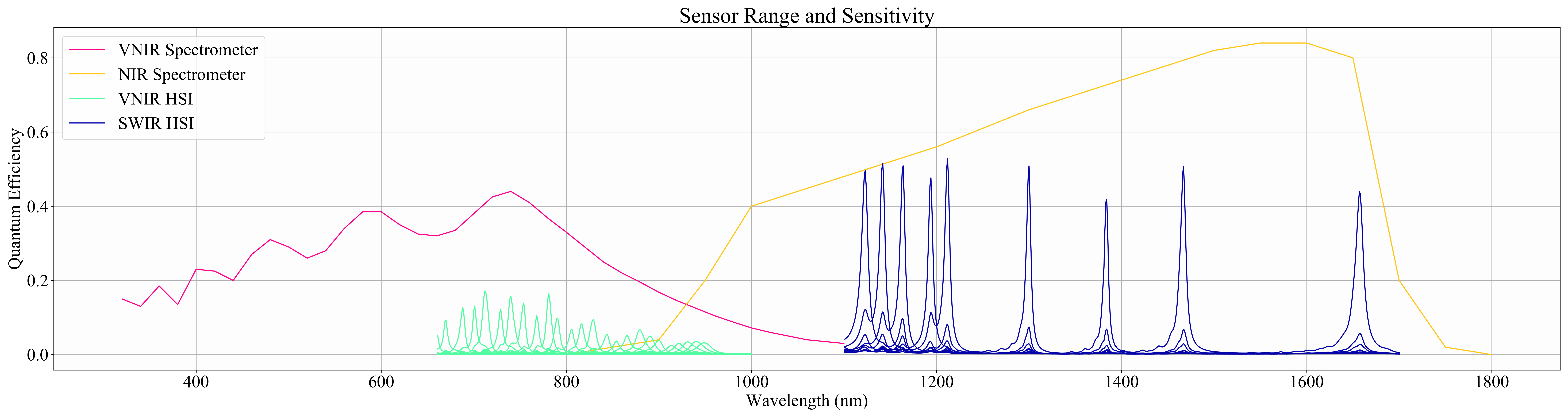} 
    \caption{Quantum efficiencies of each of the spectrum sensing devices used in these experiments. \textit{Note:} the Fabry–Pérot filters allow a spectral response typically characterized by a primary transmission peak, followed by a smaller secondary peak. This smaller feature is corrected for in the hypercube demosaicing operation.}
    \label{fig:spectral_sensitivities}
     \vspace{-1.25em}
\end{figure*}

\section{System Architecture}
\label{sec:system}
\subsection{Imaging System}
For this research we integrated two snapshot hyperspectral cameras onto a reconfigurable platform. Both sensors are manufactured by IMEC. The VNIR camera captures light from 660-900 in a 5$\times$5 Fabry–Pérot filters placed in front of a silicon-based photodetector element. When corrected, these filters produce a data cube with 24 spectral bands. Similarly, the IMEC SWIR camera captures wavelength information from 1100-1700 nm using a 3$\times$3 filter array. These filters are placed on an uncooled Indium Gallium Arsenide (InGaAs) photodetector. Unlike the VNIR camera, which has filters evenly spaced across the sensitive spectral range, the SWIR camera concentrates its bands in the lower range of the wavelength spectrum, where there are more prevalent spectral absorption features. Combined, our hyperspectral sensing solution features 33 channels, covering 1100 nm of the electromagnetic spectrum. The cameras are housed inside a 3D printed Onyx housing weatherized with epoxy. An uncoated Gorilla Glass shielding glass allows minimal perturbation of the light entering the lens.

The hyperspectral cameras are coaligned with a 5 megapixel RGB machine vision camera (Allied Vision). This system provides a high-resolution spatial reference containing the hyperspectral cameras' full field of view.

The hyperspectral cameras have a primary FOV of $25^{\circ}$. Each is calibrated using a precision checkerboard target board. The individual bands of the camera are then radially undistorted. The combined hyperspectral datacube is generated by calculating projective transforms through the checkerboard images, using bands that reveal the highest contrast between the board's squares. The combined datacube dimension is $1012 \times 1666 \times 33$; each cube is registered with the RGB camera.

\subsection{Point Spectrometer}
The protective water-proof computer housing also contains two point spectrometers (Ibsen). The Pebble VIS-NIR, with a silicon detector, is sensitive between 500 and 1100 nm with 256 spectral pixels; the Pebble NIR, with an uncooled InGaAs detector, is sensitive from 950 to 1700 nm with 128 pixels.  The two spectrometers have an overlap in the spectral signature range, which is advantageous because of the decreased quantum efficiency of the VIS-NIR spectrometer at wavelengths greater than 950 nm. As evidenced by Fig.~\ref{fig:spectral_sensitivities}, we leverage the greater efficiency of the InGaAs detector, and truncate the VIS-NIR device at wavelengths less than 950 nm. Together, the spectrometers cover a greater spectral range than the hyperspectral cameras and with an increased sensitivity. The InGaAs sensors are an order of magnitude less sensitive than the silicon-based devices, resulting in integration times that are appropriately larger.

The point spectrometers are coupled to low OH (Hydroxl) group fiber optic cables (Thor Labs). This allows the spectrometers to remain inside the weatherized housing, while still sensing light from the outside. The fiber optic cables are connected to an SMA fitting, which holds the end ferrules offset 4 cm above a 99\% Spectralon white reference target (Labsphere). The spectrometers measure a white reference signal data under the current illumination conditions, both natural and artificial. This reference signal will be used to dynamically generate reflectance calibrations in future work.

\subsection{Messaging and Computation}
Snapshot hyperspectral datacubes produced by the system are $\approx$20 megabytes. At its maximum operating rate, the system can generate nearly 1 gigabyte of data per second. To expedite data processing, the onboard compute system leverages an Intel Core i7 processor with 3 terabytes of SSD storage. 

The Robot Operating System \cite{quigley2009ros} is used to synchronize data collection from all the onboard systems. The three cameras are time-synchronized, so the RGB, VNIR, and SWIR images all correspond to the same scene at a rate of 10 Hz. All custom ROS compatible drivers, data structures, and algorithms are implemented in Python and C++ and are included as an open-source resource in the project repository. ROS enables time-synchronized data from the camera system and point spectrometers to be acquired, even when device drivers operate at different frequencies. Spectrometer messages contain time stamps, wavelength values, and raw digital counts, in addition to optional metadata fields, such as ambient humidity and device temperature. Hyperspectral data are transmitted as flattened 1-D arrays, with the dimensions of the 3-D cube, as well as meta-data fields for the central wavelengths, quantum efficiencies, and full-width half-maximum values. These message structures are extensible to other manufacturers, with examples available in the project repository.
\begin{figure*}[t]
    \vspace{0.5em}
    \includegraphics[width=\linewidth]{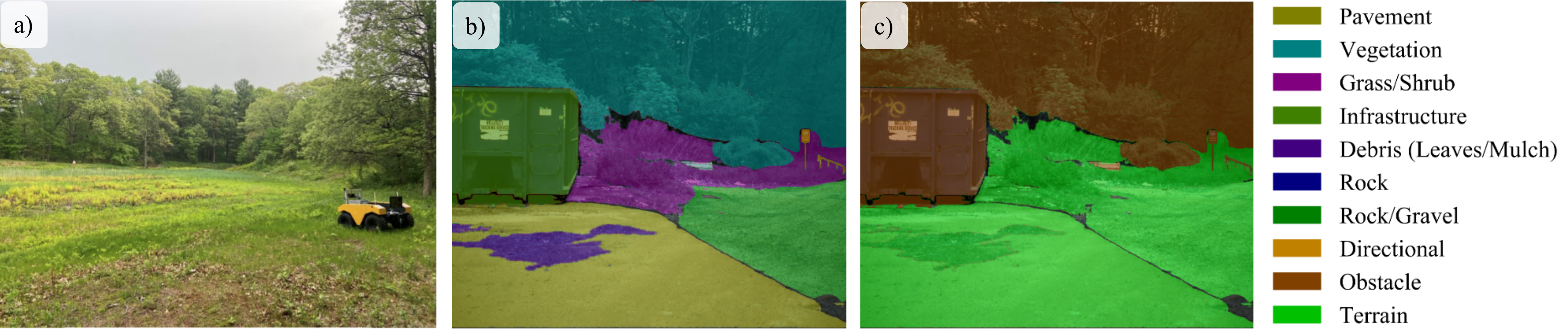} 
    \caption{a) Warthog collection vehicle traversing grass environment with dense surrounding vegetation. b) Ground-truth ATLAS segmentation labels overlayed on high-resolution RGB image. c) Reduced order labels for drivability assessments.}
    \label{fig:data_collection}
     \vspace{-1.5em}
\end{figure*}
\begin{figure}[b!]
    \vspace{-0.5em}
    \includegraphics[width=\linewidth]{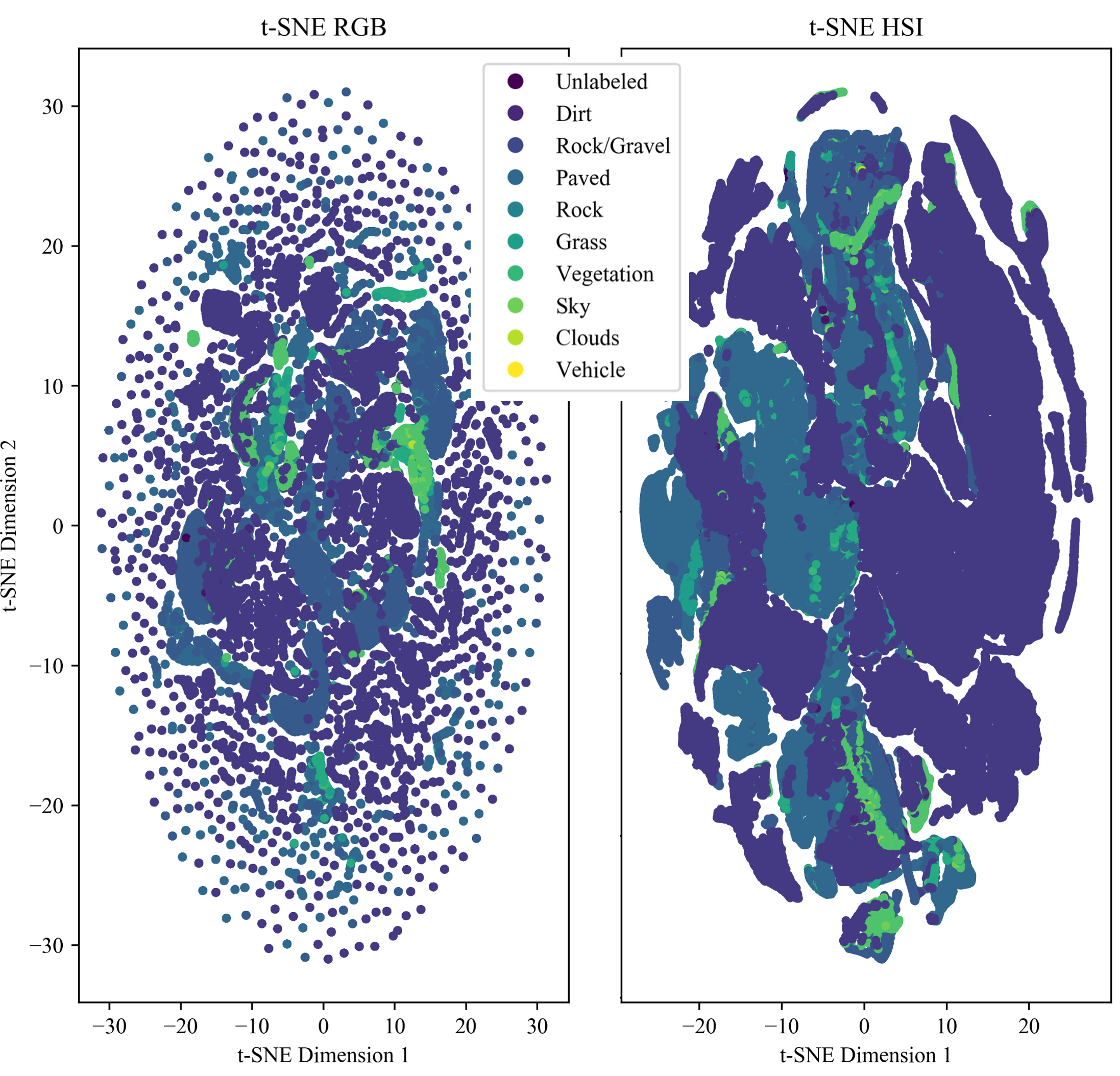}
    \caption{t-SNE two-dimensional embedding visualization of images of selected dataset image from RGB data (left) and HSI data (right).}
    \label{fig:separability_analysis}
\end{figure}
\section{HYPER DRIVE DATASET}
\label{sec:dataset}
\subsection{Dataset Construction}
We conducted a field data collection at the autonomous systems track on the campus of Olin College. The compute and sensor systems were mounted atop a Warthog unmanned ground vehicle (Clearpath Robotics). The Warthog's large ground clearance and tire treads make it an ideal candidate for mobility in both on-road and off-road terrain. 

Data were collected over two days including late afternoon, during the setting sun, early morning sunrise, and midday. This temporal variation increases the variation of solar illumination and image intensity. The vehicle was driven on 4.0 kilometers of paved roads, hiking trails, and dirt roads. We downsampled the data collection to 1 Hz to ensure scene differentiation between samples. The published dataset contains the following structure:

\begin{itemize}
    \item ROS bag file containing compressed time synchronized datastream with all following datatypes \texttt{(.bag)} 
    \item RGB image file registered to datacube \texttt{(.png)}
    \item Hyperspectral datacube compressed in 3-D array \texttt{(.npz)} \item White reference spectra from spectrometers \texttt{(.npy)}
    \item Image segmentation masks \texttt{(.png)}
\end{itemize}
\subsection{Annotations}

Previous hyperspectral datasets suffer from a lack of consistent labels, making it difficult to compare the performance of classification algorithms between datasets. We address this problem by adopting the All-Terrain Labelset for Autonomous System (ATLAS) \cite{smith2022atlas}. ATLAS provides an extensible, hierarchical ontology to generate fine-grain or coarse labels for off-road vehicle data. Each datacube has a set of labels for the whole scene: \textit{ biome, time of day, season, weather\}}. Additionally, instance labels mark the presence of specific features in the image. At the highest level, these categorizations include: \textit{\{landscape, vegetation, animal, person, obstacle, atmospheric\}}. Each of these labels can be decomposed into more specific classes or simplified to binary labels such \textit{obstacle} or \textit{landscape}. Fig.~\ref{fig:data_collection} shows images from the dataset overlayed with ATLAS labels.

As part of the initial data release, we provide 12,874 datacubes and RGB images collected from the various data collections. 500 of these images have been finely labeled with segmentation masks. To the best of the authors' knowledge, this is the largest and most diverse vehicle-centric hyperspectral dataset and the first to include shortwave infrared information. Table~\ref{tab:dataset_statistics} contains an extracted breakdown of the full label information as a function of the hierarchical classes.

\begin{table}[t]
\caption{Class Structure Statistics in HYPER DRIVE Dataset}
\vspace{-0.5em}
\label{tab:dataset_statistics}
\centering
\footnotesize
\setlength\tabcolsep{4 pt} 
\begin{tabular}{c|c|c|c}
\toprule
 Level \#1 Label & Level \# 2 Label & \# Segments & \# Images \\
\midrule
\multirow{4}{*}{Path} & Dirt & 198 & 144\\
 & Rock/Gravel & 303 & 213\\
 & Paved & 143 & 116\\
 & Concrete & 117 & 92\\
\multirow{3}{*}{Vegetation} & Ground Cover & 806 & 464\\
& Bush/Tree & 795 & 503\\
& Leaves/Mulch & 233 & 158\\
\multirow{3}{*}{Obstacle} & Vehicle & 92 & 68\\
& Infrastructure & 241 & 181\\
& Road Signage & 127 & 98\\
Person & - & 15 & 15\\
    \bottomrule
    \end{tabular}
    \vspace{-1.50em}
\end{table}
\section{DISCUSSION}
As a motivating example for why this dataset is beneficial in robotic terrain analysis, we consider the inter-class differences of the data. Fig.~\ref{fig:separability_analysis} shows a t-SNE \cite{van2008visualizing} separability analysis conducted on the data. t-SNE attempts to find a two-dimensional embedding for the natively high-dimensional hyperspectral datacube. The plot on the left shows the separability from the RGB color space alone. The plot on the right is generated from the hyperspectral datacube. 

From these embeddings, there is a clearer decision boundary between the dominant classes in the HSI t-SNE embedding. We also observe that there are more distinct groupings through the RGB data. There is still a significant amount of overlap between the classes in this two-feature representation, which is to be expected given the large number of classes (10) present in these data. The RGB plot contains a significant number of outliers in the clusters, especially in the ``dirt'' class. These tighter clusters suggest there are natural distinctions amongst the classes that semantic segmentation networks will exploit to generate more accurate classifications.
\section{CONCLUSION}
\label{sec:conclusion}
In this work, we presented a novel system architecture for collecting and associating snapshot hyperspectral data from a moving vehicle. We release a large dataset of VIS-SWIR images that encompass operating conditions as seen from a mobile robot according to the ATLAS ontology. The project also contains open-source software to integrate HSI into robotics applications. Future iterations of this system will generate fully normalized datacubes without imaging a camera white-reference image by predicting the unobserved white reference from the spectrometers' ambient spectra. We hope the introduction of a ROS framework for hyperspectral data and dissemination of our dataset will encourage further research on the applicability of HSI to other unstructured environments.

\bibliographystyle{IEEEbib}
\bibliography{strings}

\end{document}